\documentclass{article}
\usepackage{graphicx}
\usepackage{epsfig}
\usepackage{iclr2020_conference,times}
\usepackage{multirow}

\usepackage{amsmath,amsfonts,bm}

\def\eqref#1{equation~\ref{#1}}

\def\1{\bm{1}}

\DeclareMathAlphabet{\mathsfit}{\encodingdefault}{\sfdefault}{m}{sl}
\SetMathAlphabet{\mathsfit}{bold}{\encodingdefault}{\sfdefault}{bx}{n}

\usepackage{hyperref}
\usepackage{url}

\title{On the Benefits of Models with Perceptually-Aligned Gradients}

\author{%
  Gunjan Aggarwal\thanks{Authors contributed equally} \\
  Adobe Inc, Noida, India\\
  \texttt{guaggarw@adobe.com} \\
   \And
   Abhishek Sinha$^{*}$\thanks{work done while author was working at Adobe, India} \\
   Stanford University \\
   \texttt{a7b23@stanford.edu}
   \And
   Nupur Kumari$^{*}$ \\
   Adobe Inc, Noida, India\\
  \texttt{nupkumar@adobe.com} \\
  \And
   Mayank Singh$^{*}$ \\
   Adobe Inc, Noida, India\\
  \texttt{msingh@adobe.com} \\\
   }

\iclrfinalcopy 
\begin{document}

\maketitle

\begin{abstract}
Adversarial robust models have been shown to learn more robust and interpretable features than standard trained models. As shown in [\cite{tsipras2018robustness}],  such robust models inherit useful interpretable properties where the gradient aligns perceptually well with images, and adding a large targeted adversarial perturbation leads to an image resembling the target class. We perform experiments to show that interpretable and perceptually aligned gradients are present even in models that do not show high robustness to adversarial attacks. Specifically, we perform adversarial training with attack for different max-perturbation bound. Adversarial training with low max-perturbation bound results in models that have interpretable features with only slight drop in performance over clean samples. In this paper, we leverage models with interpretable perceptually-aligned features and show that adversarial training with low max-perturbation bound can improve the performance of models for zero-shot and weakly supervised localization tasks.

\end{abstract}

\section{Introduction}
Adversarial examples are crafted by adding visually imperceptible perturbation to clean inputs that leads to a significant change in the network's prediction [\cite{szegedy2013intriguing, goodfellow2014explaining}]. 
Adversarially trained classifiers are optimized using min-max objective to achieve high accuracy on adversarially-perturbed training examples [\cite{madry2017towards}]. Recently, [\cite{tsipras2018robustness, single2019imagemadry}] demonstrated that adversarial trained robust networks have gradients that are perceptually-aligned such that updating an image in the direction of the gradient to maximize the score of target class changes the image to posses visual features of the target class. Also, [\cite{lipton2019smoothing}] extended the perceptually meaningful property to randomized smoothing robust models and argued that the perceptually-aligned gradients might be a general property of robust models.

In this paper, we show that even adversarially trained models on low max-perturbation bound have perceptually-aligned interpretable features. Although these models don't show high adversarial robustness but have comparable test accuracy to standard trained models unlike adversarially trained models on high max-perturbation bound. We explore the advantages of these interpretable features on downstream tasks and show improvements on zero-shot and weakly supervised localization tasks.


\section{Experiments And Results}
In this section, we start with a brief description of the datasets and the model architecture used, following up with details of the experiment.

\vspace{-4pt}
\paragraph{Datasets} 
We perform experiments on five standard datasets, USPS [\cite{uspsdata}], MNIST [\cite{mnist_dataset}], CIFAR-10[\cite{krizhevsky2010cifar}], SVHN[\cite{netzer2011reading}] and Restricted Imagenet[\cite{tsipras2018robustness}]. 
{MNIST} consists of grayscale handwritten images of digits of size $28*28$. We use the network architecture as described in \cite{attack2017pgd}.
{USPS}, like MNIST, has handwritten digits for 10 different classes. However, the image size is $16\times16$. MNIST and USPS are subject to different data distributions and thus are often used to check zero-shot or transfer learning performance of models. {CIFAR-10} consists of RGB images of size $32\times32$ for 10 different classes. {SVHN} contains RGB images of streetview house numbers for 10 different classes. {Restricted ImageNet} consists of a subset of ImageNet classes which have been grouped into 9 different classes that has RGB images of dimension $224\times224$. We perform experiments over WRN-28-10 model [\cite{wrn2016Zagoruyko}] for SVHN, CIFAR-10 and over ResNet50 [\cite{he2016deep}] for Restricted ImageNet dataset.

\paragraph{Methodology} We perform adversarial training (AT) using either FSGM [\cite{goodfellow2014explaining, fast2020fgsm}] or PGD [\cite{madry2017towards}]. We found similar results on the mentioned downstream tasks for FGSM and PGD adversarially trained models. 
\paragraph{Qualitative evaluation}
We perform adversarial training (AT) for CIFAR-10 and SVHN over WRN-28-10 model under $l\infty$ norm constraint, for a range of maximum perturbation bound($\epsilon$). Different models were trained for $\epsilon \in [1.0/255.0, 2.0/255.0, 4.0/255.0, 8.0/255.0]$ with a batch size of $128$ for $30,000$ and $5,000$ steps for CIFAR-10 and SVHN respectively. Let us denote the adversarially trained model with $l\infty$ perturbation bound($\epsilon$) as AT-$\epsilon$ and standard trained models as $\textit{Natural}$ that trains using the clean samples only.

The robustness of  different models against a strong adversarial attack, PGD attack, for various values of $\epsilon$ over CIFAR-10 dataset have been mentioned in table \ref{table:CIFAR}. We find that models trained against small $\epsilon$ adversarial perturbations do not show high robustness against strong adversarial attacks.

\begin{table}
\centering
\scalebox{0.8}{
\begin{tabular}{|c|c|c|c|c|c|c|c|}
\hline
{Model} & \multicolumn{3}{c|}{{Adversarial Accuracy (10 step PGD attack)}}\\
& $\epsilon = 0.0/255.0$ & $\epsilon = 4.0/255.0$ & $\epsilon = 8.0/255.0$\\
 \hline \hline
\textit{Natural} & 90.88 & 0.01 & 0.00\\
\textit{AT-1}  & 90.95 & 34.33 & 4.53\\
\textit{AT-2}  & 90.88 & 50.76 & 14.41\\
\textit{AT-4}  & 89.79  & 61.28 & 27.30\\
\textit{AT-8}  & 87.25 & 73.99 & 47.37\\
\hline
\end{tabular}
}
\caption{Adversarial Robustness of different models on CIFAR-10}
\label{table:CIFAR}
\end{table}

\begin{figure*}[t]
\centering
\scalebox{1.0}{
    \includegraphics[width=2.45in,height=2.45in]{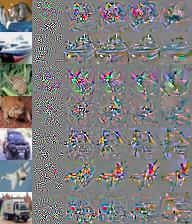} 
    \includegraphics[width=2.45in,height=2.45in]{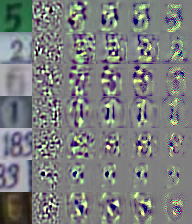}
   } 
    \caption{\footnotesize{Visualization of image level gradients for CIFAR-10 and SVHN dataset. Column 1 denotes the original image. Columns 2-7 denote the gradient from \textit{Natural, AT-1, AT-2, AT-4, AT-8} models respectively}}
    \label{fig:grad}
\end{figure*}

\begin{figure*}[t]
\centering
\scalebox{1.0}{
    \includegraphics[width=2.45in,height=2.45in]{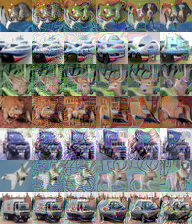} 
    \includegraphics[width=2.45in,height=2.45in]{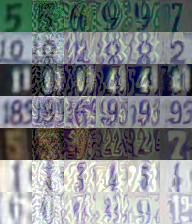}
 }   
    \caption{\footnotesize{High $\epsilon$ adversarial images. Column 1 denotes the original image. Columns 2-7 denote high $\epsilon$ adversarial images from \textit{Natural, AT-1, AT-2, AT-4, AT-8} models respectively}}
    \label{fig:high-eps}
\end{figure*}

The image level gradients for all the different models as done in \cite{tsipras2018robustness} have been visualized in figure \ref{fig:grad}. As can be seen from the figure, that while the gradients of the \textit{natural} model looks like a noisy version of the original image, the gradients of the model trained via adversarial training for different $\epsilon$ align well with the original image. This is valid even for the adversarial trained models with low value of $\epsilon$.

\begin{figure*}[t]
\centering
\scalebox{0.2}{
\includegraphics{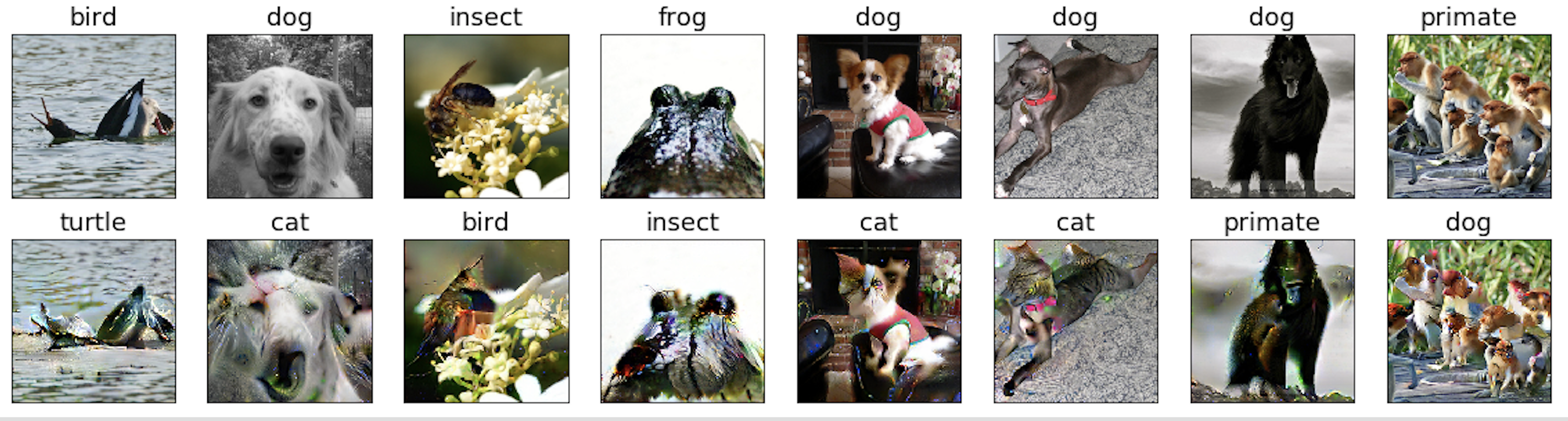}
}
\caption{\footnotesize{Example images generated by large $\epsilon$ untargeted $l_2$ attack using ResNet-50 model trained by adversarial training with $\epsilon=2$ on Restricted ImageNet. Top row denotes the original images while the bottom row denotes the adversarial images.}}
\label{fig:attack}
\end{figure*}

We also perform non-targeted adversarial attack at a high value of $\epsilon = 32.0/255.0$. We use PGD attack, with a learning rate of $1.6/255.0$ and step size of $50$ to compute the attack. Again we find that, while such an attack induces random noise to the \textit{Natural} model, for the other models such adversarial images often resemble those belonging to different classes. The results are shown in figure \ref{fig:high-eps}. We can see cat changing to dog, frog changing to deer and truck changing to car for the different models over the CIFAR-10 dataset.
We repeat this experiment over the Restricted ImageNet dataset too. Figure \ref{fig:attack} shows some of the images generated from a model trained via adversarial training for $\epsilon=2$ using ResNet50 model.

Interestingly both of the above phenomenons were shown to be property of adversarial robust models in \cite{tsipras2018robustness}. In contrast, we show that not so robust models, trained via adversarial training for small values of $\epsilon$ also exhibit such properties.


\vspace{-8pt}
\paragraph{Zero shot learning} Models that seem to learn more interpretable features should also perform better against shifts in data distributions. We show the performance of adversarial trained models for two zero-shot tasks, SVHN$\to$MNIST and MNIST$\to$USPS. For both of these tasks, the model was trained over the source dataset and evaluated over the validation set of both source and target datasets. We define \textit{sourceAcc.} and \textit{targetAcc.} as the validation accuracies over the source and target datasets respectively.


\textbf{SVHN$\to$MNIST} We perform adversarial training for $\epsilon$ $\in$ [1.0/255.0, 2.0/255.0, 4.0/255.0, 8.0/255.0] over the SVHN dataset. Again \textit{AT-1,  AT-2, AT-4, AT-8} refer to the corresponding adversarial trained models. As a baseline we also perform training of models with gaussian noise of varying strength added to the inputs, as done in [\cite{lipton2019smoothing}]. The corresponding models are referred to as \textit{N-1, N-2, N-4, N-8}.
The source and target validation accuracies have been mentioned in table \ref{table:SVHN}.

\begin{table}[t]
    \begin{minipage}{.5\linewidth}
      \centering
        \begin{tabular}{|c|c|c|c|c|c|c|c|}
            \hline
            \textbf{Model} & \textbf{sourceAcc.} & \textbf{targetAcc.} \\
             \hline \hline
            \textit{Natural} & 94.72 & 53.98 \\
            \textit{AT-1/N-1}  & \textbf{95.40}/95.36 & 75.99/74.65 \\
            \textit{AT-2/N-2}  & 94.73/94.93 & 75.73/72.14 \\
            \textit{AT-4/N-4}  & 93.31/94.21 & \textbf{77.26}/74.30  \\
            \textit{AT-8/N-8}  & 90.02/90.68 & 69.25/58.20 \\
            \hline
        \end{tabular}
        \caption{\footnotesize{Zero-shot accuracy for SVHN$\to$MNIST}}
        \label{table:SVHN}
    \end{minipage} 
    \begin{minipage}{.5\linewidth}
    \centering
      \begin{tabular}{|c|c|c|c|c|c|c|c|}
        \hline
        \textbf{Model} & \textbf{sourceAcc.} & \textbf{targetAcc.} \\
         \hline \hline
        \textit{Natural} & 99.17 & 75.14 \\
        \textit{AT-0.05/N-0.05}  & \textbf{99.37}/98.71 & 81.81/80.32 \\
        \textit{AT-0.1/N-0.1}  & 99.23/99.1 & \textbf{82.41}/81.81 \\
        \textit{AT-0.2/N-0.2}  & 99.04/99.2  & 77.68/79.92 \\
        \textit{AT-0.3/N-0.3}  & 98.4/99.19 & 73.99/81.22 \\
        \hline
        \end{tabular}
      \caption{\footnotesize{Zero-shot accuracy for MNIST$\to$USPS}}
      \label{table:MNIST}
    \end{minipage}
\end{table}

\begin{figure*}[t]
\centering
    \begin{minipage}{.3\textwidth}
      \includegraphics[width=\linewidth]{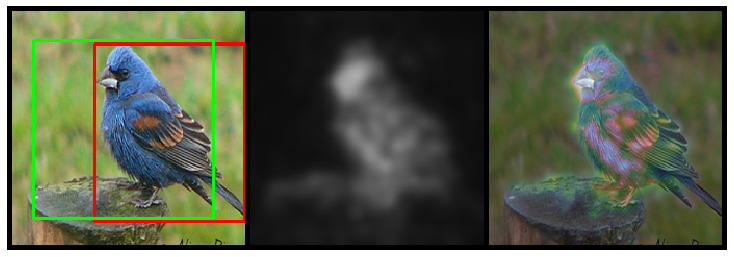}
    \end{minipage}
    \begin{minipage}{.3\textwidth}
      \includegraphics[width=\linewidth]{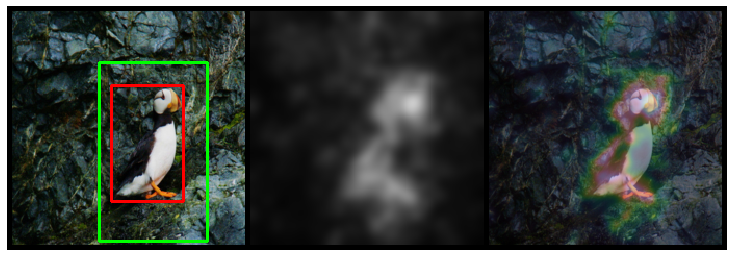}
    \end{minipage}
    \begin{minipage}{.3\textwidth}
      \includegraphics[width=\linewidth]{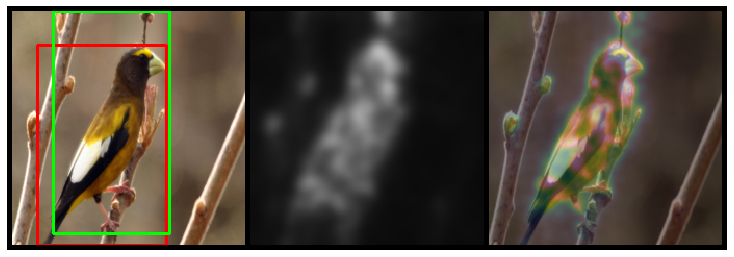}
    \end{minipage}\\
    \begin{minipage}{.3\textwidth}
      \includegraphics[width=\linewidth]{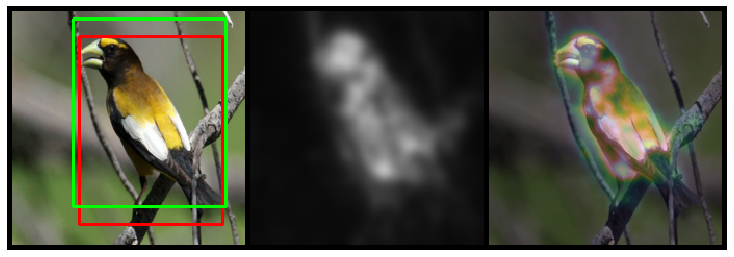}
    \end{minipage}
    \begin{minipage}{.3\textwidth}
      \includegraphics[width=\linewidth]{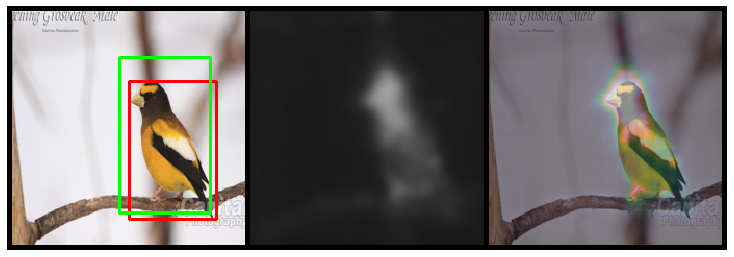}
    \end{minipage}
    \begin{minipage}{.3\textwidth}
      \includegraphics[width=\linewidth]{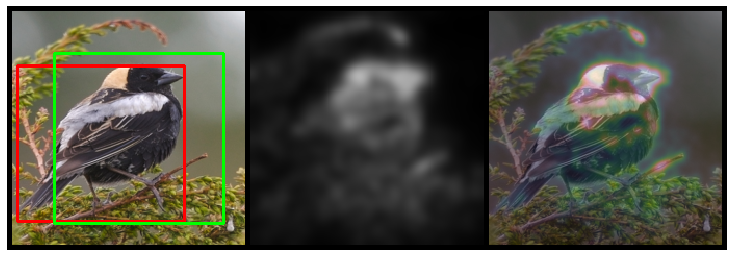}
    \end{minipage}\\
    \caption{\footnotesize{Examples of estimated bounding box and heatmap by adversarially ResNet50 model on randomly chosen images of CUB dataset; Red bounding box is ground truth and green bounding box corresponds to the estimated box}}
    \label{fig:loc_our}
\end{figure*}
 
\begin{table}
\centering
\scalebox{0.675}{
\begin{tabular}{|c|c|c|c|c|}
\hline
\textbf{Model} & Method & \multicolumn{2}{c|}{\textbf{WSOL}} & Top-1 Acc \\
& & GT-Known Loc & Top-1 Loc  & \\ \hline 
ResNet50-SE &  ADL\citep{adl} & - & \textbf{62.29} & 80.34 \\ \hline
\multirow{4}{*}{ResNet50} &  Natural & 60.37 & 50.0 & 81.12 \\ \cline{2-5}
&  AT-0.5 & 77.93 & \underline{61.03} & 78.69 \\ \cline{2-5}
&  AT-1& 74.55 & 55.89 & 75.88 \\ \cline{2-5}
&  AT-2& 69.93 & 50.10 & 70.02 \\ 
\hline

\end{tabular}}
\caption{\footnotesize{Weakly Supervised Localization on CUB dataset. Bold text refers to the best Top-1 Loc and underline denotes the second best.}}
\label{table:weakloc}
\end{table}

\textbf{MNIST$\to$USPS} We perform adversarial training for $\epsilon$ $\in$ [0.05,0,1,0.2,0.3] over the MNIST dataset and then evaluate the performance of the trained model over the USPS dataset. The results have been mentioned in table \ref{table:MNIST}.

For both zero-shot tasks, we find that the adversarial trained models for low $\epsilon$ often achieve higher performance over the target dataset than the naturally trained models. Further, the performance of AT models for low $\epsilon$ often exceeds the AT models trained on high $\epsilon$

\paragraph{Weakly Supervised Object Localization (WSOL)} aims to localize the object in the given image through weak supervisory signals like image labels instead of rich annotation of bounding boxes. Most of the prior techniques \citep{adl,wsol_prev_2,wsol_prev_1} exploit CAM \citep{attr2016cam} image interpretation technique for predicting the bounding box of the object using a trained model. We hypothesize that models having perceptually aligned gradients with respect to image and high test accuracy should prove useful in this domain. We show empirically that adversarially trained model with low max-perturbation bound having perceptually aligned gradients performs comparable to the state-of-the-art in WSOL on CUB \citep{cub} dataset in table \ref{table:weakloc}. We report on evaluation metrics used in \cite{adl} i.e. Top-1 classification accuracy (\textit{Top-1 Acc}); Localization accuracy when ground truth is known (\textit{GT-Known Loc}), i.e when intersection over union (IoU) of estimated box and ground truth bounding box$>$0.5; Top-1 localization accuracy, i.e. when prediction is correct and IoU of bounding box$>$0.5 (\textit{Top-1 Loc}). We show some sample images and estimated bounding boxes in figure \ref{fig:loc_our}.
\hspace{-4mm}
\section{Conclusion}
We observe that adversarial training on low values of $\epsilon$ also induces features that are perceptually-aligned unlike standard trained model. While these models do not show strong robustness against adversarial attacks, but performing adversarial attack of large magnitude often generates an image belonging to a different class. We further show that such models outperform models trained naturally and AT models with high value of $\epsilon$, on two different zero-shot tasks and a weakly supervised localization task. We hope that our findings will spur research aimed at harnessing the perceptually-aligned model features to further generalize to additional downstream tasks.

\bibliography{iclr2020_conference}
\bibliographystyle{iclr2020_conference}

\appendix

\end{document}